\documentclass[final,5p,times, twocolumn]{elsarticle}

\usepackage{hyperref}
\usepackage{lineno}
\usepackage{comment}
\usepackage{amsmath}
\usepackage{float}
\usepackage{graphicx}
\usepackage{svg}
\usepackage{pdfpages} 
\modulolinenumbers[5]
\usepackage{pgfplots}
\usepackage{amssymb}
\usepackage{multirow}
\usepackage{ gensymb }
\usepackage{multicol}
\usepackage{makecell}
\usepackage{caption}
\usepackage{fixltx2e}

\makeatletter
\g@addto@macro{\UrlBreaks}{\UrlOrds}
\makeatother

\usepackage{subfig}
\usepackage{caption}
\journal{arXiv}

\begin{document}
\begin{frontmatter}

\title{ \huge Broad-UNet: Multi-scale feature learning for nowcasting tasks}

\author{Jesús García Fernández} 
\ead{j.garciafernandez@student.maastrichtuniversity.nl}

\author{Siamak Mehrkanoon \corref{cor1}}
\ead{siamak.mehrkanoon@maastrichtuniversity.nl}

\cortext[cor1]{Corresponding author}

\address{Department of Data Science and Knowledge Engineering, Maastricht University, The Netherlands}

\begin{abstract}
Weather nowcasting consists of predicting meteorological components in the short term at high spatial resolutions. Due to its influence in many human activities, accurate nowcasting has recently gained plenty of attention. In this paper, we treat the nowcasting problem as an image-to-image translation problem using satellite imagery. We introduce Broad-UNet, a novel architecture based on the core UNet model, to efficiently address this problem. In particular, the proposed Broad-UNet is equipped with asymmetric parallel convolutions as well as Atrous Spatial Pyramid Pooling (ASPP) module. In this way, The the Broad-UNet model learns more complex patterns by combining multi-scale features while using fewer parameters than the core UNet model. The proposed model is applied on two different nowcasting tasks, i.e. precipitation maps and cloud cover nowcasting. The obtained numerical results show that the introduced Broad-UNet model performs more accurate predictions compared to the other examined architectures.

\end{abstract} 

\begin{keyword}
Satellite imagery \sep Precipitation forecasting \sep cloud cover forecasting \sep Deep learning \sep Convolutional neural network \sep U-Net
\end{keyword}
\end{frontmatter}

\section{Introduction}
Weather forecasting is an essential task that has a great influence on humans daily life and activities. Industries such as agriculture \cite{cogato2019extreme}, mining \cite{ivanov2019weather} and construction \cite{senouci2018impact} rely on the weather forecasts to make decisions and thus unexpected climatological events may result in large economic losses. Similarly, accurate weather forecasts improve safety on flights and roads and help us foresee potential natural disasters.

Due to its importance, precipitation nowcasting is becoming an increasingly popular research topic. This term refers to the problem of forecasting precipitation in the near future at high spatial resolutions. It is usually performed through satellite imagery and many different approaches have been proposed for this problem. Classical nowcasting approaches mainly focus on two methods: Numerical Weather Prediction (NWP) \cite{sun2014use} and extrapolation based techniques, such as Optical Flow (OF) \cite{woo2017operational}. 
NWP methods simulate the underlying physics of the atmosphere and ocean to generate predictions, so they require a vast amount of computational resources. In contrast, optical flow based methods identify and predict how objects move through a sequence of images. But they are unable to represent the dynamics behind them.
In recent years, the massive amount of existing data has aroused research interest in data driven machine learning techniques for nowcasting \cite{shi2015convolutional, holmstrom2016machine, grover2015deep}. By taking advantage of available historical data, data-driven based approaches have shown better performance than classical ones in many forecasting tasks \cite{faloutsos2019classical}. Furthermore, while classical machine learning techniques rely on handcrafted features and domain knowledge, deep learning techniques automatize the extraction of those features. Recent advances in deep learning have shown promising results in diverse research areas such as neuroscience, biomedical signal analysis, weather forecasting and dynamical systems, among others \cite{webb2018deep,mehrkanoon2018deep,mehrkanoon2019deep,mehrkanoon2015learning,mehrkanoon2019cross, gamboa2017deep, salman2015weather, coban2018neuro, coban2013context}.
Convolutional Neural Networks (CNNs) are the most popular algorithms used in computer vision \cite{voulodimos2018deep}, achieving the state-of-the-art in various tasks \cite{lu2007survey, voulodimos2018deep, goel2020state}. CNN architectures, such as AlexNet \cite{krizhevsky2017imagenet}, ResNet \cite{he2016deep} and InceptionNet \cite{szegedy2015going}, to name a few, mainly consist of the combination of convolutional and pooling layers. They are outstanding at classification, identification and recognition tasks. 
Among other architectures, autoencoders have emerged as one of the most powerful approaches in both supervised \cite{zhang2019light, berthomier2020cloud, fernandez2020deep} and unsupervised learning \cite{baldi2012autoencoders, lample2017unsupervised, chung2016audio} with the UNet \cite{ronneberger2015u} being one of the most versatile architectures.
The UNet architecture was first proposed for medical image segmentation, but it has been employed in various domains \cite{trebing2021smaat, tao2017background, fernandez2020deep}. It consists of a contracting path, to extract features, and an expanding path, to reconstruct a segmented image, with a set of residual connection between them to enable precise localization. In our previous work \cite{fernandez2020deep}, we introduced various extended versions of the UNet for weather forecasting problem. In this paper, we further extend the best performing model in that work \cite{fernandez2020deep}, i.e. the AsymmIncepRes3DDR-UNet. In particular, motivated by the results of \cite{chen2017deeplab}, we augment the AsymmIncepRes3DDR-UNet's feature extraction capacity by incorporating an Atrous Spatial Pyramidal Pooling module (ASPP) \cite{chen2017deeplab} in the bottleneck of the network. The ASPP module works in line with the existing building blocks of our network (Multi-scale feature convolutional block), extracting multi-scale features in parallel and combining them. Therefore, unlike the original UNet, the proposed model is designed to capture multi-scale information. In addition, it keeps the temporal dimension unchanged along the encoder path and then reduces it before being concatenated with the output of every level in the decoder path. As a result, it can efficiently learn a mapping between 3-dimensional input data and 2-dimensional output data. 
Furthermore, we apply a kernel factorization in most of the convolutional operations of the model, resulting in a significant reduction in the total number of parameters compared to the original UNet while having improved performance. These techniques are explained in detail in the subsequent sections. 
We further present an analysis of this multi-scale features extraction and the enhancement provided by the ASPP module. We show its versatility by applying it to two different nowcasting tasks, i.e. precipitation nowcasting and cloud cover nowcasting. In the precipitation nowcasting task, the model performs a regression of every pixel. In the case of cloud cover nowcasting, the model classifies each pixel as containing clouds or not. In addition, we directly compare the proposed model with the model introduced in \cite{trebing2021smaat}, a variation of UNet architecture that relies on depthwise-separable convolutions and includes a CBAM attention module \cite{woo2018cbam} at each level. While the model in \cite{trebing2021smaat} approximates the performance of the original UNet with a significantly reduced number of parameters, our model outperforms the original UNet with a reduced number of parameters.


\section{Related work}
Traditionally, optical flow based models are the most popular techniques among classical methods for precipitation nowcasting tasks  \cite{bowler2004development, li2018subpixel}. However, machine learning and deep learning based approaches are dominating this field of research in recent years. Due to the vast amount of available satellite imagery, powerful deep neural networks based models are suitable candidates that can be used to address various problems existing in this field. In particular, CNN based architectures have show their great ability to handle 2D and 3D images. Thanks to the versatility of CNN's, nowcasting problems can be tackled in different fashions. For instance, the authors in \cite{ayzel2019all} and \cite{agrawal2019machine} treated the multiple time-steps as multiple channels in the network. In this way, they could apply a simple 2D-CNN to perform the predictions. Additionally, the authors in \cite{shi2017deep} treated the multiple time-steps as depth in the samples. Thus they can apply a 3D-CNN and approximate more complex functions. As it has been shown in \cite{lebedev2019precipitation,agrawal2019machine,trebing2021smaat}, among the used CNN architectures, UNet is more suitable for this task, due to its autoencoder-like architecture and ability to tackle image-to-image translation problems. 

In addition to CNN's, Recurrent Neural Networks (RNN's) have proved to be a robust approach. However, these architectures struggle to work with images but can capture long-range dependencies, an ability that CNN's can only partially achieve with the addition of attention mechanisms, such as self-attention \cite{vaswani2017attention}. In \cite{shi2015convolutional}, the authors introduce an architecture that combines both CNN's and RNN's strengths. They extend the fully connected LSTM (FC-LSTM) with convolutional structures, obtaining the Convolutional LSTM network (ConvLSTM). As a result, the proposed model captures spatiotemporal correlations better than the FC-LSTM model. The authors in \cite{shi2017deep} introduce the Trajectory GRU (TrajGRU) model as an extension of the ConvLSTM. This architecture keeps the advantages of the previous model and also learns the location-variant structure of the recurrent connections, showing superior performance than the other models compared. Nevertheless, these RNN models have not been directly compared with the UNet in nowcasting tasks. \\
The authors in \cite{berthomier2020cloud} make a comparison among different types of models for cloud cover nowcasting. In \cite{berthomier2020cloud}, the models under assessment are various versions of CNN's, RNN's, LSTM and UNet. The authors showed that the UNet model is the best performing model for the given cloud cover nowcasting task.

\section{Proposed model}\label{sec:proposedmodels}
In this section, we introduce our Broad-UNet model. First, different elements that are used for building the network are presented. The complete architecture is then explained. 

\begin{figure}[!htbp]
    \centering
    \includegraphics[width=0.96\columnwidth]{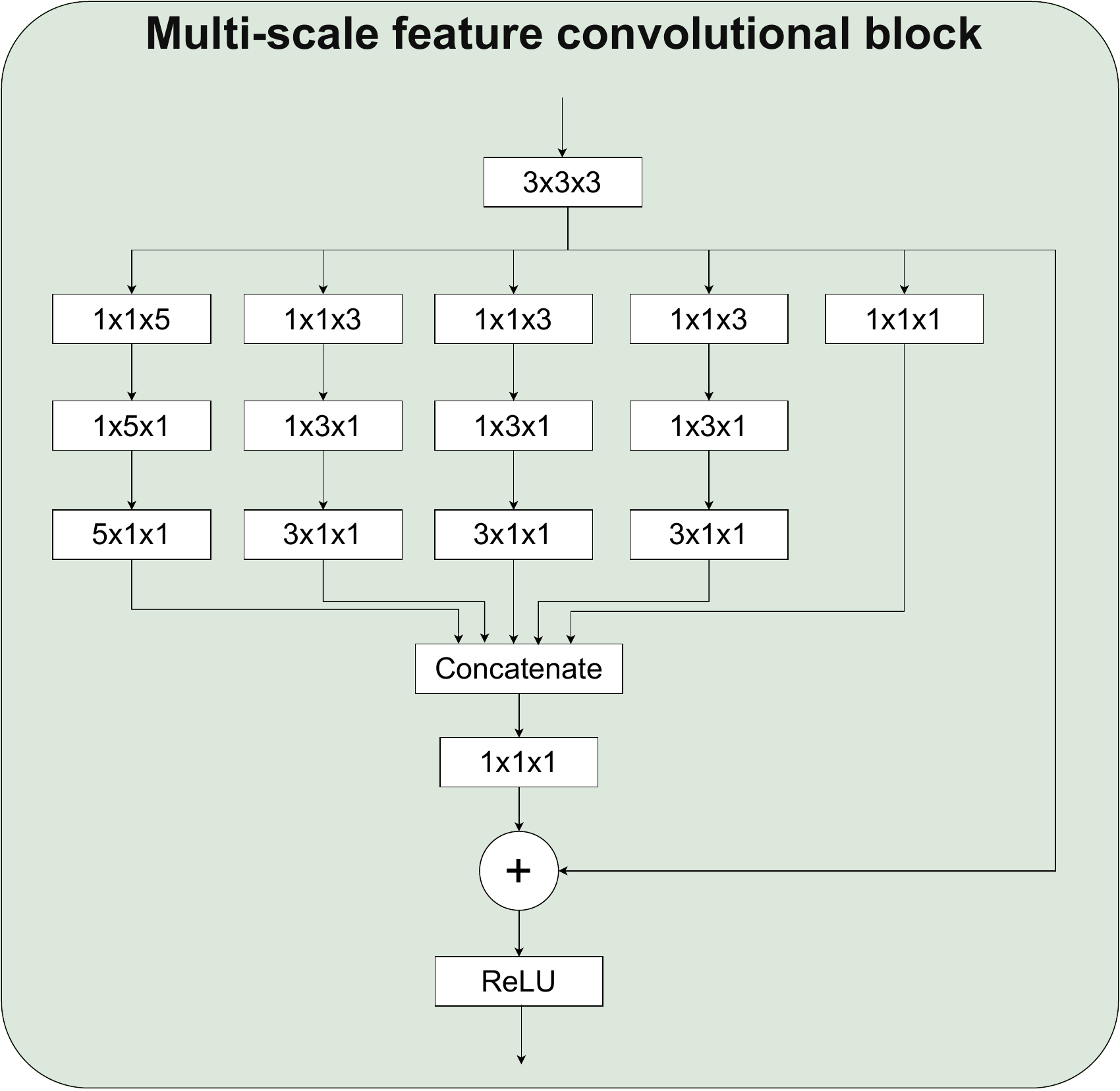}
    \caption{Multi-scale feature convolutional block. Convolutions with different kernels are performed in parallel to extract features at different scales. A residual connection also keeps some unmodified information.}
    \label{fig:convBlock}
\end{figure}

\subsection{Multi-scale feature convolutional block}

Motivated by the goal of extracting features at different scales, the model contains a block consisting of parallel arms as shown in Fig \ref{fig:convBlock}. This block serves as the core building block of our network. Within this block, the data forks into parallel branches of convolutions with different kernel sizes, after going through an initial convolution. A $3\times3\times3$ convolution is followed by a set of parallel convolutions with $1\times1\times1$, $3\times3\times3$ and $5\times5\times5$ kernel sizes. The outputs of the different branches are then concatenated and merged with a $1\times1\times1$ convolution. Additionally, inspired by the results found in \cite{szegedy2016inception}, we keep some information intact alongside the parallel branches with a residual connection. Lastly, the output of the block is rectified with a ReLU activation function. To reduce the large number of features resulting from these branches, we factorize the convolutions as suggested in \cite{yang2019asymmetric}. That means a convolution $N \times N\times N$ decomposes into the three consecutive $1\times1\times N$, $1\times N\times1$ and $N\times1\times1$ convolutions. Hence, this sequence is an approximation of the original convolution with fewer parameters.


\begin{figure}[!htbp]
    \centering
    \includegraphics[width=0.96\columnwidth]{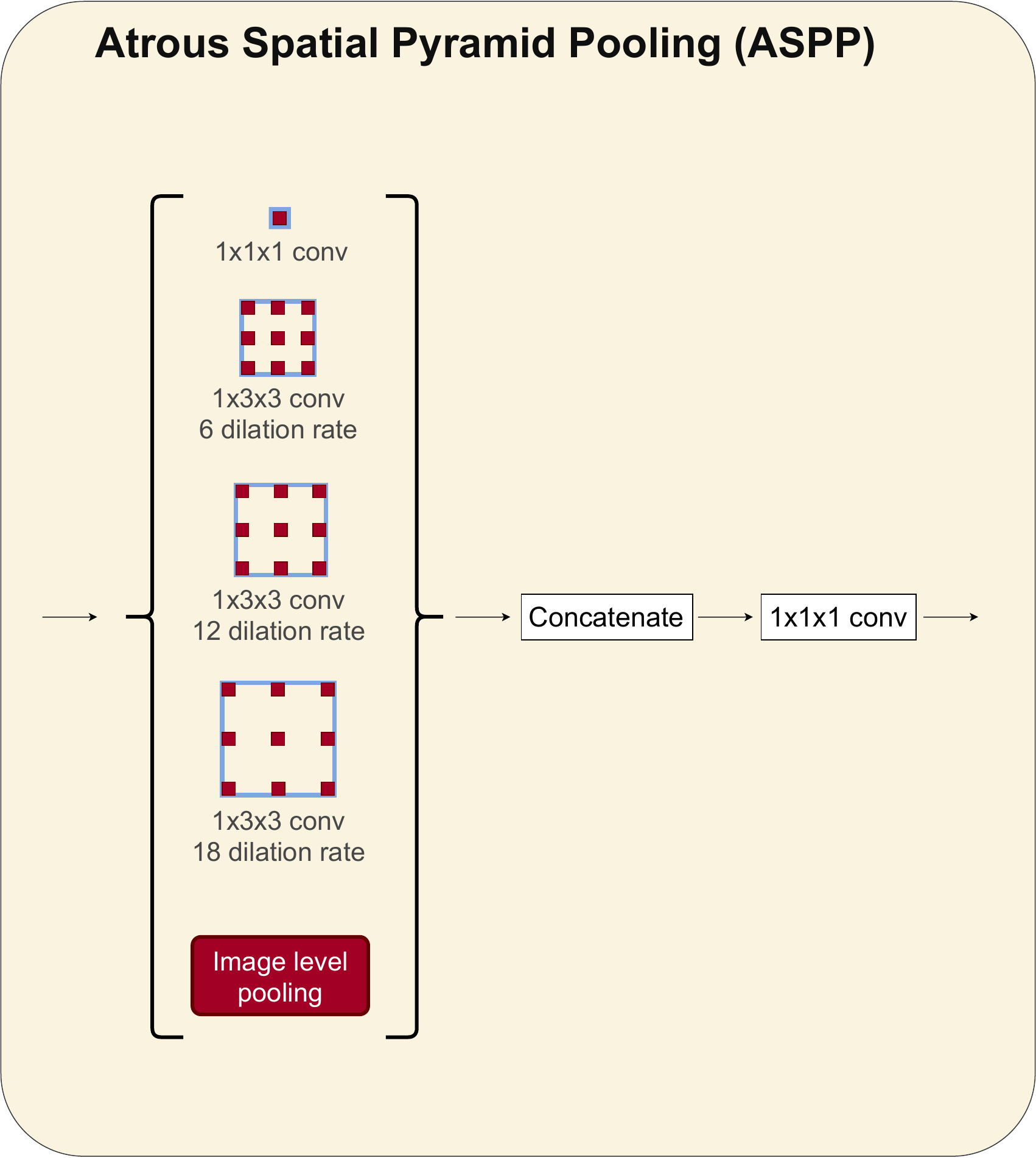}
    \caption{Atrous Spatial Pyramidal Pooling (ASPP) block. Different dilation rates allow the network to extract multi-scale information. Due to the kernel shapes ($1\times N\times N$) and the image-level pooling mechanism, only spatial information is extracted.}
    \label{fig:ASPP}
\end{figure}

\subsection{Atrous Spatial Pyramid Pooling (ASPP)}

Atrous Spatial Pyramid Pooling (ASPP), is a mechanism used to capture multi-scale information. It consists of parallel branches of convolutions, similar to the convolutional block presented above. However, instead of using different kernel sizes, the same kernel is chosen with an increasing dilation rate (6, 12 and 18). In this kind of convolutions, the filter is upsampled by inserting zeros between successive values. As a result, they employ a larger field of view, without experiencing an explosion in the number of parameters. Further, it only extracts information in the spatial dimensions by applying a 2-dimensional filter (shape $1\times N\times N$). 
In addition, ASPP incorporates one branch to extract image-levels features, allowing to capture global context information. Here, we implement it by applying a global average pooling, and subsequent reshaping and upsampling back. The obtained extracted features are then concatenated and combined with a $1\times1\times1$ convolution. The scheme of this mechanism is shown in Fig. \ref{fig:ASPP}.

\subsection{Broad-UNet}
\begin{figure*}[!htbp]
    \centering
    \hspace{-0.5cm}
    \includegraphics[scale=0.29]{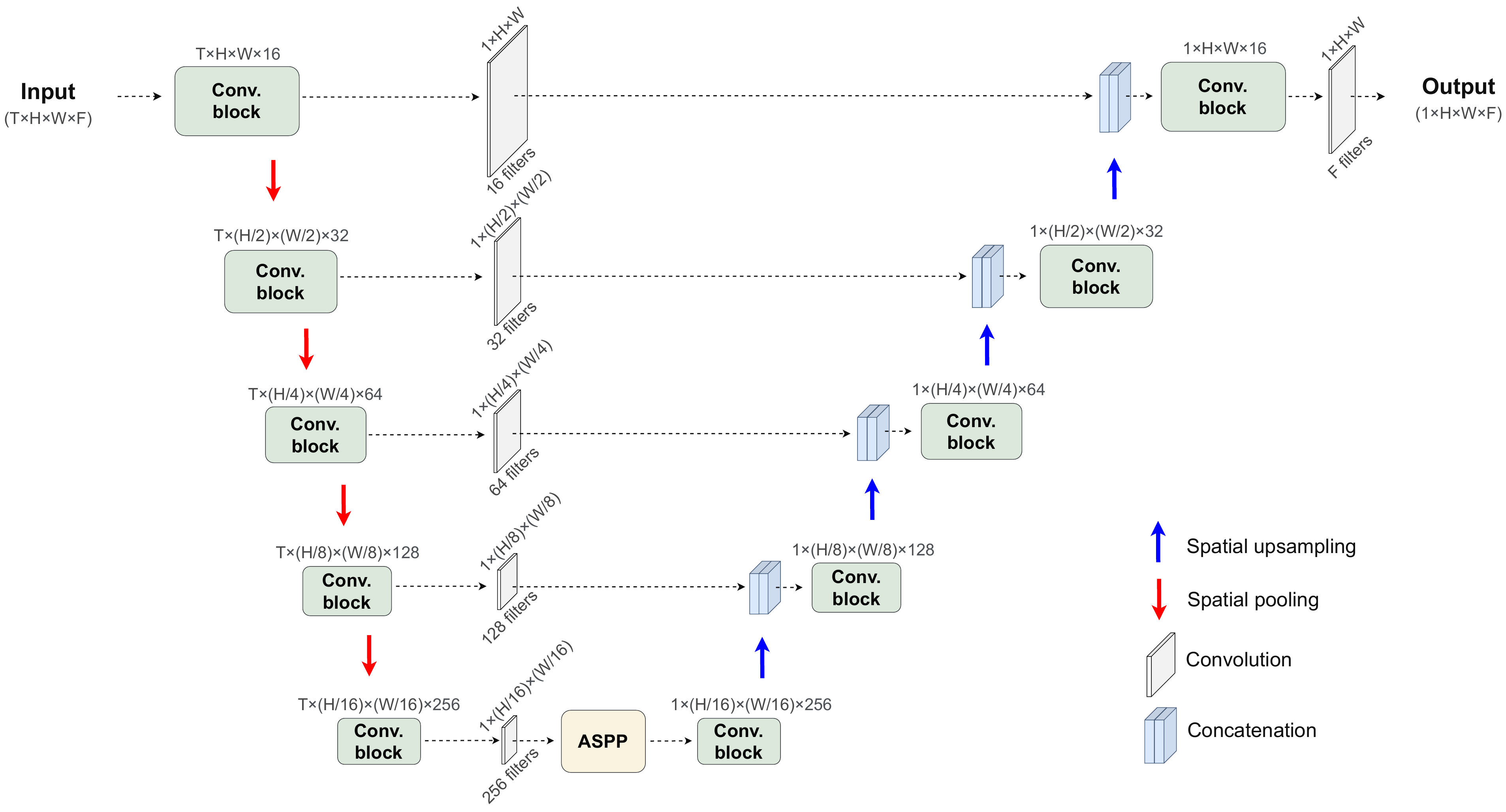}
    \caption{Complete architecture of Broad-UNet. The Multi-scale feature convolutional block is displayed as \textit{Conv. Block} for simplicity. The annotation over these blocks describes the output dimension, where \textit{T} represents the time-steps (lags), \textit{H} and \textit{W} represent the height and width of each image and \textit{F} the number of features or elements predicting.}
    \label{fig:architecture}
\end{figure*}
Thanks to the effectiveness of UNet architecture in solving image-to-image mapping tasks, it is chosen to serve as the basis to construct our model. UNet core model which was originally proposed for medical image segmentation tasks, adopts an autoencoder structure. While the encoder part extracts features from the input image, the decoder part performs classification on each pixel to reconstruct the segmented output. Plus, a set of residual connections between both parts allows a precise localization in the output image. 
Differently, our proposed Broad-UNet manipulates 3-dimensional data in the encoder and 2-dimensional data in the decoder. Thus, we can input several time-steps in the first dimension, and it outputs only one time-step in the same dimension. Multi-scale feature convolutional blocks are alternated with pooling operations in the encoder, resulting in five levels. 
The pooling is only performed in the spatial dimensions (2nd and 3rd) and implemented with a Max Pooling layer. In this way, the temporal dimension of the data remains unchanged. Then, the decoder follows a similar structure. It alternates multi-scale feature convolutional blocks and upsampling in the spatial dimensions. Additionally, we incorporate extra convolutions in the connections between different levels of the encoder and decoder. These intermediate convolutional operations aim to reduce the temporal dimension
from \textit{T} time-steps to 1.

To extend the multi-scale feature learning process, we combine the convolutional blocks with the ASPP module. It is placed in the bottleneck of the network, where the data has a highly abstract representation. In this way, we allow the network to capture more information from this representation without using larger kernels and more computational resources. Also, dropout is included in the bottleneck to force the network to learn a more sparse representation of the data and avoid possible overfitting. As a result, the network input is of shape $T \times H\times W \times F$ and output is of shape $1 \times H\times W \times F$, where \textit{T} is the number of time-steps (lags), \textit{H} and \textit{W} are the height and width of the images, and \textit{F} is the number of features or elements, which we consider as channels in our network. Here, the convolutions to reduce the temporal dimension have a kernel size $1 \times T\times T$ with valid padding. In addition, the use of asymmetric convolutions drastically reduces the total number of parameters of the network. While the number of parameters is $\sim$28 million using regular kernels $N \times N\times N$, the number of parameters after factorizing the convolutions into $1 \times 1\times N$, $1 \times N\times 1$ and $N \times 1\times 1$ is $\sim$11 million. The complete architecture of the model can be found in Fig. \ref{fig:architecture}. Furthermore, a comparison in the number of learnable parameters among different UNet based models examined in this paper is shown in Table \ref{fig:parameters}.

\begin{table}[!htbp]
    \centering
    \caption{Comparison between the number of learnable parameters in different UNet based models examined in this paper.}
     \begin{tabular}{|c c|} 
     \hline 
     \textbf{Model} & \textbf{Number of parameters}\\ [0.5ex] 
     \hline\hline
     UNet & $\sim$17M\\ 
     \hline
     SmaAt-UNet & $\sim$4M \\
     \hline
     AsymmIncepRes3DDR-UNet & $\sim$9.5M \\
     \hline
     Broad-UNet & $\sim$11M \\
     \hline
    \end{tabular}
\label{fig:parameters}
\end{table}

\section{Data description and preprocessing}\label{sec:datadescription}
To assess the performance of our model, we apply it to two different datasets. Both of them consist of satellite images and are intended to tackle weather nowcasting problems. The first one includes precipitation maps, in which the value of each pixel shows the amount of rainfall in that region. The second one consists of cloud cover maps, in which the pixel values are binary and indicate whether there is a cloud or not in that region. Here, we recreate the same samples as in \cite{trebing2021smaat} for the first dataset, and as in \cite{berthomier2020cloud} for the second dataset. In this way, we can make a fair comparison with the results obtained in those research works. For reproducibility purposes, all our models and scripts are available on Github \footnote{\url{https://github.com/jesusgf96/Broad-UNet}}. Also, the datasets and pre-trained models are available upon request.

\subsection{Precipitation maps dataset} \label{subsec:precipdata}
The first dataset, provided by the Royal Netherlands Meteorological Institute (Koninklijk Nederlands Meteorologisch Instituut, KNMI) \cite{KNMI}, includes rainfall measurements from two Dutch radar stations (De Bilt and Den Helder). These measurements are in the shape of images. The images cover the region of the Netherlands and neighbouring countries, spanning four years in 5-minutes intervals. To train and validate the models, we use data from the years 2016-2018 (80\% train/ 20\% validation), and the data from 2019 is used as test set. 

\begin{figure}[!htbp]
\hspace{-0.1cm}
\begin{minipage}[b]{0.495\linewidth}
\centering
\includegraphics[width=1\textwidth]{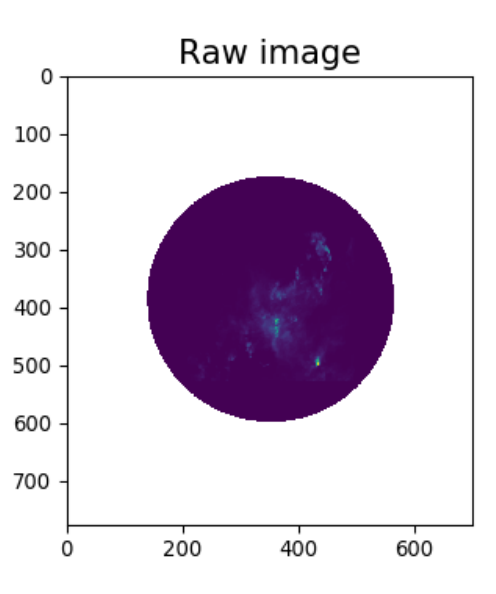}
\end{minipage}
\begin{minipage}[b]{0.495\linewidth}
\centering
\includegraphics[width=1.04\textwidth]{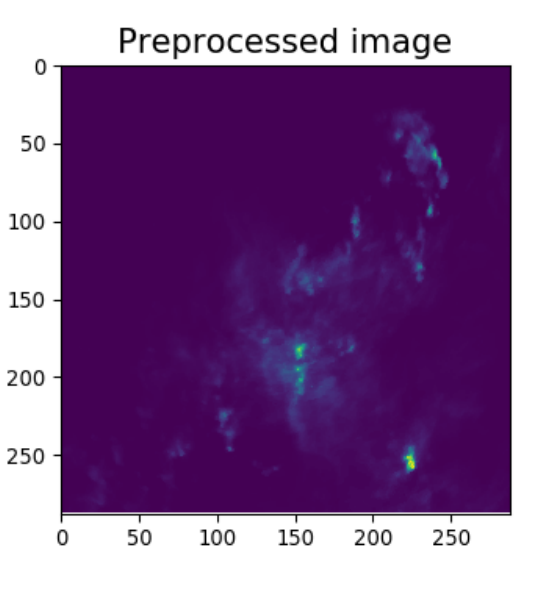}
\end{minipage}
\caption{Example of precipitation map from the first dataset, before and after applying the preprocessing.}
\label{fig:precipitationMap}
\end{figure}

The values of each pixel represent the accumulated amount of rainfall in the last five minutes. That means that a value \textit{n} represents $n \times 10^{-2} $ mm of rain per square kilometre. The resolution of the images is $765 \times 700$, and the measured region is circle-shaped with a large margin. Following the lines of \cite{trebing2021smaat}, we cropped the central squared area with size $288 \times 288$, as shown in Fig. \ref{fig:precipitationMap}.

Moreover, there is a high imbalance between pixels with rain and no rain, with plenty of images lacking raining pixels. Therefore, as in \cite{trebing2021smaat}, we filter the dataset choosing only the images with at least 50\% of pixels containing any amount of rain. This dataset is then used to create the training/validation/test samples. Additionally, we create a second dataset filtering the images with at least $20\%$ of pixels containing any amount of rain. From this second dataset, we use only the test set. Therefore, it serves as a way of testing our trained models under different conditions. We also normalize both datasets by dividing them by the highest value in the training set.

\subsection{Cloud cover dataset} \label{subsec:clouddata}
The second dataset is the "Geostationary Nowcasting Cloud Type" classification product \cite{clouds} from the European Organisation for the Exploitation of Meteorological Satellites (EUMETSTAT). It is composed of satellite images of the whole globe at longitude $0$ degrees, taken every 15 mins. The resulting size of the images in the dataset is $3712 \times 3712$, spanning the years 2017-2018. For multi-comparison purposes, we generate two different dataset from this data. In the first dataset, we follow the lines of \cite{berthomier2020cloud} and use data from 2017 and the first semester of 2018 as training set. Then the data from the second semester of 2018 is used for both validation and test. On the contrary, as in \cite{trebing2021smaat}, we use data from 2017 and the first semester of 2018 for train and validate our models (80\% train / 20\% validation). The data from the second semester of 2018 is thus used only for test. We use data from 2017 and the first semester of 2018 to train the models. To validate and test the models, we use data from the second semester of 2018. 
In this data, every pixel can have $15$ different values (1: Cloud-free land, 2: Cloud-free sea, 3: Snow over land, 4: Sea ice, 5: Very low clouds, 6: Low clouds, 7: Mid-level clouds, 8: High opaque clouds, 9: Very high opaque clouds, 10: Fractional clouds, 11: High semitransparent thin clouds, 12: High semitransparent meanly thick clouds, 13: High semitransparent thick clouds, 14: High semitransparent above low or medium clouds, 15: High semitransparent above snow/ice). However, following the lines of \cite{berthomier2020cloud}, we aim to perform a classification between cloud or no-cloud. Therefore, we group the labels from 1 to 4 into 0 (no-cloud) and the labels from 5 to 15 into 1 (cloud). Also, we crop the images according to the boundaries of France: [51.896, 41.104, -5.842, 9.842] (upper latitude, lower latitude, left longitude, right longitude). Then we apply a transformation to obtain a suitable projection and reshape the resulting image to $256\times256$ pixels. Fig. \ref{fig:cloudsImages}, displays an example of the described pre-processing steps.

\begin{figure}[!htbp]
\hspace{-0.1cm}
\begin{minipage}[b]{0.495\linewidth}
\centering
\includegraphics[width=1.035\textwidth]{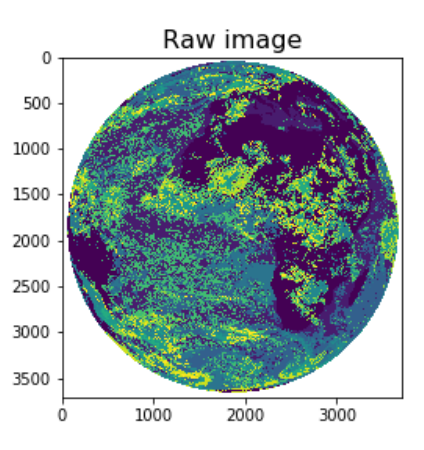}
\end{minipage}
\begin{minipage}[b]{0.495\linewidth}
\centering
\includegraphics[width=1\textwidth]{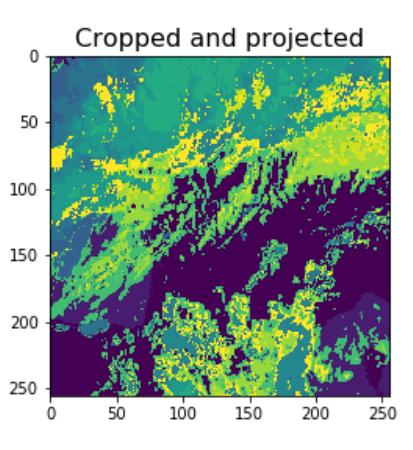}
\end{minipage}
\centering
\includegraphics[width=0.25\textwidth]{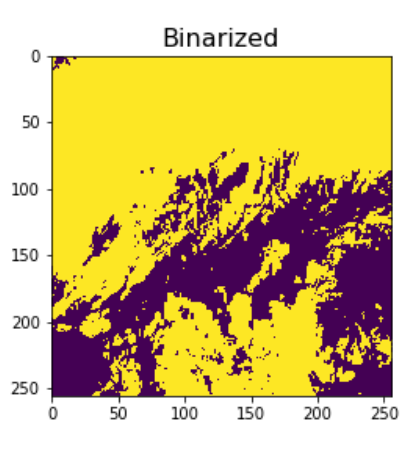}
\caption{Example of an image from the cloud cover dataset, before and after applying the different steps of the preprocessing.}
\label{fig:cloudsImages}
\end{figure}

\section{Experimental setup and evaluation}
In order to have a fair comparison with the results obtained in \cite{trebing2021smaat} and \cite{berthomier2020cloud} for both datasets, we reproduce the same experimental setups. The data is arranged in such a way that the resulting input is a four-dimensional array $\mathcal{I} \in \mathbb{R}^{T \times H \times W \times F}$, where $T$ is the number of lags or previous time-steps, which corresponds to the time dimension. $H$ and $W$ refer to the size of the image and make up the spatial dimensions. The last element $F$ corresponds to the predicted features, which in both cases is $1$. We use TensorFlow to implement our models and train and evaluate them on the given datasets. 
The hyperparameters of models are tuned and the optimal ones are empirically found and used. 


\subsection{Precipitation maps nowcasting} \label{subsec:precipdataEval}
As for the precipitation maps dataset, we apply the preprocessing and split the dataset as described in section \ref{subsec:precipdata}. We aim to predict a precipitation map $30$ minutes ahead or considering that the images are generated five minutes apart, six time-steps ahead. The number of lags, previous time-steps, is set to 12 which was emprically found to be the best one among othre tested lag values. The height and width of the images are 288 and 288, and the number of features in the input is 1, i.e. the precipitation maps. Therefore, the inputs of the model has the shape (12, 288, 288, 1), and output data has the shape (1, 288, 288, 1). 

In this nowcasting task, we perform a regression of every pixel. 
Mean Squared Error (MSE) is used as the loss function and Adam optimizer to optimize it, with an initial learning rate of 0.0001. The batch size and the dropout rate are set to 2 and 0.5, respectively. We also implemented a checkpoint callback to monitor the validation loss. 
Thus the best performing model on the validation set is saved. We use MSE as the main metric to assess the performance of the model in this dataset. Furthermore, we also include additional metrics such as accuracy, precision and recall. Following the lines of \cite{trebing2021smaat}, in order to calculate these new metrics, we first create a binarized mask of the image, according to a threshold. This threshold is the mean value of the training set from the 50\% of rain pixel dataset. Hence, any value equal or over the threshold is replaced by 1, and any value under it is replaced by 0.

\subsection{Cloud cover nowcasting} \label{subsec:evalclouddata}
Regarding the cloud cover dataset, we preprocess the data and split the dataset as described in section \ref{subsec:clouddata}. In this case, we predict six different time-steps: from 15 minutes to 70 minutes ahead, or from 1 to 6 time-steps ahead. Due to the architecture of our network, we train six different model. Thus each model predicts a different time-step. Here, the number of lag is set to 4, the height and width of the images are 256 and 256, and the number of input features is again 1, the cloud cover map. That means that the model receives input data with the shape (4, 256, 256, 1), and outputs data with the shape (1, 256, 256, 1). In this task, we perform binary classification of every pixel. 
The binary cross-entropy is used as the loss function in this case. We use Adam optimizer with an initial learning rate of 0.001. The batch size and the dropout rate are set to 8 and 0.5, respectively. Similarly, we implemented a checkpoint callback to monitor the validation loss. 
Thus the best performing model on the validation set is saved. Following the lines of \cite{berthomier2020cloud}, here we also use MSE as the metric to assess the performance of the model. First, we calculate the MSE between the ground truth and the raw prediction as the main metric. In this case, the values between 0 and 1 in the predictions indicate the probability of cloud occurrence in that region. In addition, we binarize the prediction of the network with a threshold of 0.5 to generate a second assessment with the MSE metric.
We also include additional metrics, i.e. accuracy, precision and recall, to compare the performance of the Broad-UNet with the model introduced in \cite{trebing2021smaat}, which also uses UNet architecture as the basis. To calculate these new metrics, we first create a binarized mask of the image, using the value 0.5 as the threshold.

\section{Results}

\subsection{Precipitation maps nowcasting}
\begin{figure}[H]
\centering
\includegraphics[trim={0 1cm 0 1cm},clip, width=\columnwidth]{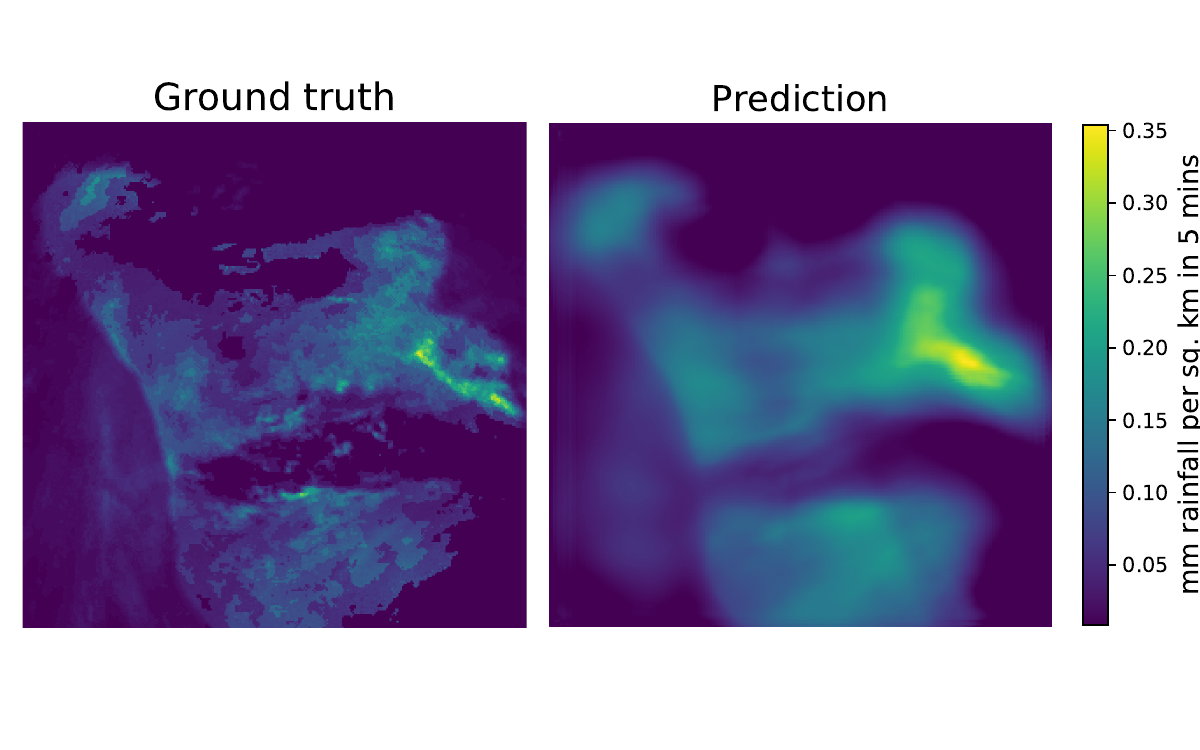}
\centering
\includegraphics[trim={0 1cm 0 1cm},clip,width=\columnwidth]{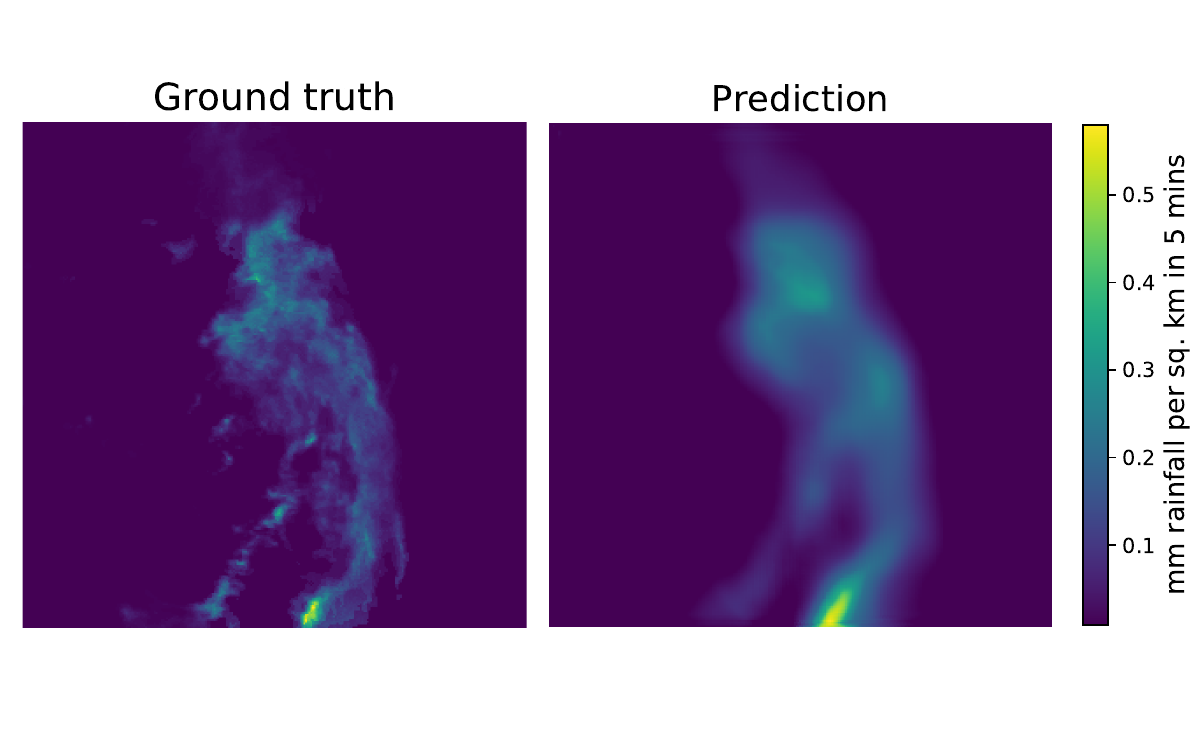}
\caption{Broad-UNet precipitation prediction examples. The images in the first row are generated with the test set from the 50\% of pixels containing rain dataset. The images in the second row are generated with the test set from the 20\% of pixels containing rain dataset.}
\label{fig:precipImages}
\end{figure}

In the precipitation maps prediction task, we compare the performance of the Broad-UNet with the persistence, a simple meteorological baseline used in forecasting, and different models over the test sets of two different datasets, i.e. 50\% of rain pixels and 20\% of rain pixels. These models are the UNet \cite{ronneberger2015u} and two variants \cite{trebing2021smaat, fernandez2020deep}. The MSE is the main metric used for this comparison, and it is calculated over the denormalized data. The additional metrics are computed over the binarized data, as described in section \ref{subsec:precipdataEval}.
The performance of different models over the first precipitation maps dataset is shown in Table \ref{tab:resultsPrecipitation50}. In the same way, the performance of the models in the second precipitation maps dataset is listed in Table \ref{tab:resultsPrecipitation20}. From the obtained results, one can observe that the Broad-UNet achieved the lowest MSE score in both datasets. 
\begin{table}[!htbp]
    \centering
    \scriptsize{
    \renewcommand{\arraystretch}{1.5}
    \caption{Test MSE and additional metrics values for the precipitation maps prediction task using the 50\% of rain pixels dataset. $\downarrow$ indicates that the optimal values are the smallest ones and $\uparrow$ indicates that the optimal values are the highest ones.}
    \label{tab:resultsPrecipitation50}
    \begin{tabular}{l l c c c}
    \Xhline{3\arrayrulewidth}
    &\multicolumn{4}{c}{\textbf{MSE 50\% of rain pixels dataset}} \\
    \Xhline{3\arrayrulewidth}
    \multirow{2}{*}{\textbf{Model}}& 
    \multirow{2}{*}{\textbf{ MSE $\downarrow$}} & 
    \multirow{2}{*}{\textbf{Accuracy $\uparrow$}} & 
    \multirow{2}{*}{\textbf{Precision $\uparrow$}} & 
    \multirow{2}{*}{\textbf{Recall $\uparrow$}}\\
     & & & & \\\Xhline{3\arrayrulewidth}
         \textbf{Persistance} & 2.48e-02 & 0.756 & 0.678 & 0.643	\\
         \textbf{UNet} & 1.22e-02 & 0.836 & 0.740 & \underline{0.855}	\\ 
         \textbf{SmaAt-UNet} & 1.22e-02 & 0.829 & 0.730 & 0.850	\\ 
         \textbf{AsymmIncepRes3DDR-UNet} & 1.11e-02 & \underline{0.858} & \underline{0.759} & 0.800	 \\ 
         \textbf{Broad-UNet} & \underline{1.08e-02} & 0.850 & 0.715 & 0.817	\\ 
        \Xhline{3\arrayrulewidth}
    \end{tabular}
    }
\end{table}
\begin{table}[!htbp]
    \centering
    \scriptsize{
    \renewcommand{\arraystretch}{1.5}
    \caption{Test MSE and additional metrics values for the precipitation maps prediction task using the 20\% of rain pixels dataset. $\downarrow$ indicates that the optimal values are the smallest ones and $\uparrow$ indicates that the optimal values are the highest ones.}
    \label{tab:resultsPrecipitation20}
    \begin{tabular}{l l c c c}
    \Xhline{3\arrayrulewidth}
    &\multicolumn{4}{c}{\textbf{MSE 20\% of rain pixels dataset}} \\
    \Xhline{3\arrayrulewidth}
    \multirow{2}{*}{\textbf{Model}}& 
    \multirow{2}{*}{\textbf{MSE $\downarrow$}} & 
    \multirow{2}{*}{\textbf{Accuracy $\uparrow$}} & 
    \multirow{2}{*}{\textbf{Precision $\uparrow$}} & 
    \multirow{2}{*}{\textbf{Recall $\uparrow$}}\\
     & & & & \\\Xhline{3\arrayrulewidth}
         \textbf{Persistance} & 2.28e-02 & 0.827 & 0.559 & 0.543	\\
         \textbf{UNet} & 1.11e-02 & 0.880 & \underline{0.666} & 0.782	\\ 
         \textbf{SmaAt-UNet} & 1.11e-02 & 0.867 & 0.626 & \underline{0.801}	\\ 
         \textbf{AsymmIncepRes3DDR-UNet} & \underline{1.02e-02} & 0.893 & 0.621 & 0.767 \\ 
         \textbf{Broad-UNet} & \underline{1.02e-02} & \underline{0.895} & 0.611 & 0.772	\\ 
        \Xhline{3\arrayrulewidth}
    \end{tabular}
    }
\end{table}
Two examples of $30$ minutes ahead prediction with the Broad-UNet are displayed in Fig. \ref{fig:precipImages}. The images on the first and second row are generated using the first and second precipitation dataset, respectively.

\subsection{Cloud cover nowcasting}

When applying the Broad-UNet to the second dataset, we compare its performance with the persistence and various models. These models are introduced and explained in \cite{berthomier2020cloud}. We perform this comparison with the results obtained from the test set of the cloud cover dataset. The used evaluation metrics are explained in section \ref{subsec:evalclouddata}.
In Fig. \ref{fig:mseClouds}, we show the MSE obtained using the ground truth and the actual prediction. 
Fig. \ref{fig:mseCloudsBin} depicts the MSE calculated with the ground truth and binarized prediction. 
From Fig. \ref{fig:mseClouds} and Fig. \ref{fig:mseCloudsBin}, one can notice that the Broad-UNet performance is superior in short-term forecasting. 
As the number of step-ahead increases, the gap between the performance of the proposed Broad-UNet and the classical UNet model decreases.
In addition, in Table \ref{tab:resultsCloudVariousMetrics}, we show the comparison between the Broad-UNet's and the model introduced in \cite{trebing2021smaat}. As in \cite{trebing2021smaat}, the metrics tabulated in this table are averaged over different time-steps (15-90 minutes ahead). From the obtained results one can observe that the Broad-UNet performs better than other compared models in three out of four used metrics.
\begin{figure}[!htbp]
\centering
\resizebox{!}{0.62\columnwidth}{
\begin{tikzpicture}
\begin{axis}[
    title={},
    xlabel={Minutes ahead},
    ylabel={MSE},
    xmin=15, xmax=90,
    ymin=0.03, ymax=0.15,
    xtick={15,30,45,60,75,90},
    ytick={0,0.05,0.1, 0.15},
    legend pos=outer north east,
    ymajorgrids=true,
    grid style=dashed,
]
\addplot[color=red,]
    coordinates {(15,0.053228)(30,0.066889)(45,0.076967)(60,0.084975)(75,0.092092)(90,0.098159)};
\addplot[color=green,]
    coordinates {(15,0.060833)(30,0.072903)(45,0.081940)(60,0.090815)(75,0.098411)(90,0.104633)};
\addplot[color=cyan,]
    coordinates {(15,0.063139)(30,0.073429)(45,0.083607)(60,0.092693)(75,0.100649)(90,0.107282)};
\addplot[color=black,]
    coordinates {(15,0.061075)(30,0.074556)(45,0.084235)(60,0.096763)(75,0.107243)(90,0.113285)};
\addplot[color=blue,]
    coordinates {(15,0.05101)(30,0.0650)(45,0.0771)(60,0.0856)(75,0.0920)(90,0.0972)};
\legend{U-Net,CNN,LSTM,RNN, Broad-UNet}
\end{axis}
\end{tikzpicture}}
\caption{Test MSE values of the different models for the cloud cover prediction task.}
\label{fig:mseClouds}
\end{figure}
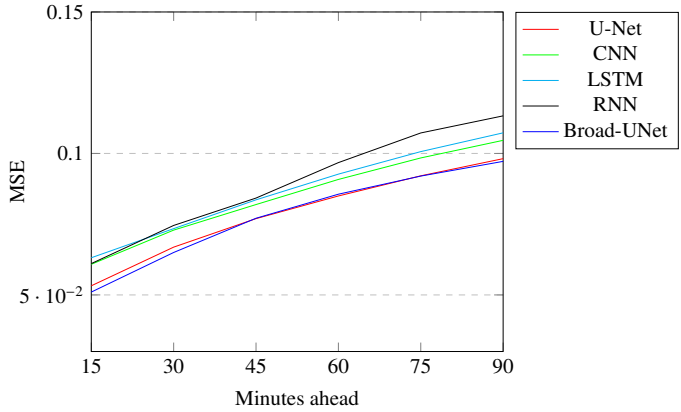
\begin{figure}[!htbp]
\centering
\resizebox{!}{0.62\columnwidth}{
\begin{tikzpicture}
\begin{axis}[
    title={},
    xlabel={Minutes ahead},
    ylabel={Binarized MSE},
    xmin=15, xmax=90,
    ymin=0.03, ymax=0.17,
    xtick={15,30,45,60,75,90},
    ytick={0,0.05,0.1,0.15},
    legend pos=outer north east,
    ymajorgrids=true,
    grid style=dashed,
]
\addplot[color=red,]
    coordinates {(15,0.074002)(30,0.093589)(45,0.108022)(60,0.119442)(75,0.129743)(90,0.138677)};
\addplot[color=green,]
    coordinates {(15,0.082896)(30,0.099692)(45,0.113601)(60,0.125846)(75,0.136501)(90,0.145644)};
\addplot[color=cyan,]
    coordinates {(15,0.087416)(30,0.102051)(45,0.116400)(60,0.128994)(75,0.140283)(90,0.149806)};
\addplot[color=black,]
    coordinates {(15,0.081668)(30,0.099434)(45,0.113222)(60,0.127803)(75,0.140324)(90,0.149110)};
\addplot[color=blue,]
    coordinates {(15,0.0691)(30,0.0911)(45,0.1080)(60,0.1202)(75,0.1298)(90,0.1376)};
\legend{U-Net,CNN,LSTM,RNN, Broad-UNet}
\end{axis}
\end{tikzpicture}}
\caption{Test MSE values of the different models for the cloud cover prediction task with binarized predictions.}
\label{fig:mseCloudsBin}
\end{figure}
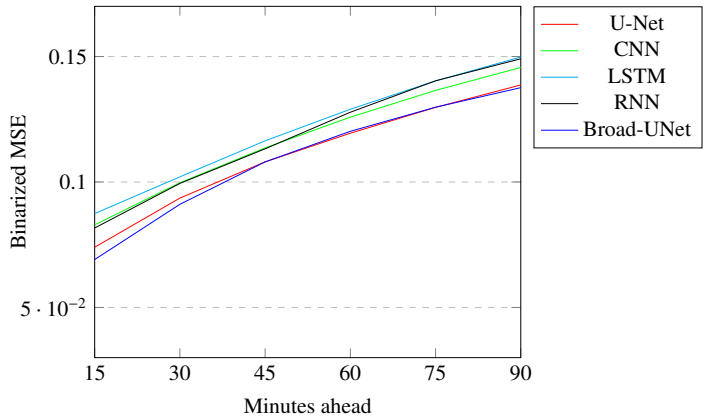
\begin{table}[!htbp]
    \centering
    \scriptsize{
    \renewcommand{\arraystretch}{1.5}
    \caption{Average of the MSE and additional metrics for the cloud cover prediction task. $\downarrow$ indicates that the optimal values are the smallest ones and $\uparrow$ indicates that the optimal values are the highest ones.}
    \label{tab:resultsCloudVariousMetrics}
    \begin{tabular}{l l c c c}
    \Xhline{3\arrayrulewidth}
    &\multicolumn{4}{c}{\textbf{Averaged test MSE cloud cover prediction}} \\
    \Xhline{3\arrayrulewidth}
    \multirow{2}{*}{\textbf{Model}}& 
    \multirow{2}{*}{\textbf{ MSE $\downarrow$}} & 
    \multirow{2}{*}{\textbf{Accuracy $\uparrow$}} & 
    \multirow{2}{*}{\textbf{Precision $\uparrow$}} & 
    \multirow{2}{*}{\textbf{Recall $\uparrow$}}\\
     & & & & \\\Xhline{3\arrayrulewidth}
         \textbf{Persistance} & 0.1491 & 0.851 & 0.849 & 0.849	\\ 
         \textbf{UNet} & 0.0785 & 0.890 & 0.895 & 0.919	\\ 
         \textbf{SmaAt-UNet} & 0.0794 & 0.889 & 0.892 & \underline{0.921}	\\ 
         \textbf{Broad-UNet} & \underline{0.0783} & \underline{0.891} & \underline{0.898} & 0.914\\ 
        \Xhline{3\arrayrulewidth}
    \end{tabular}
    }
\end{table}
In addition, two examples of the Broad-UNet's predictions are displayed in Fig. \ref{fig:cloudsImages}. Both predictions are generated with the test set of the cloud cover dataset. In Fig. \ref{fig:cloudsImages}, the images on the first and second row correspond to 30 mins and 90 mins ahead prediction, respectively.

\begin{figure}[!htbp]
\centering
\includegraphics[trim={0 0.1cm 0 0.1cm},clip, width=1.02\columnwidth]{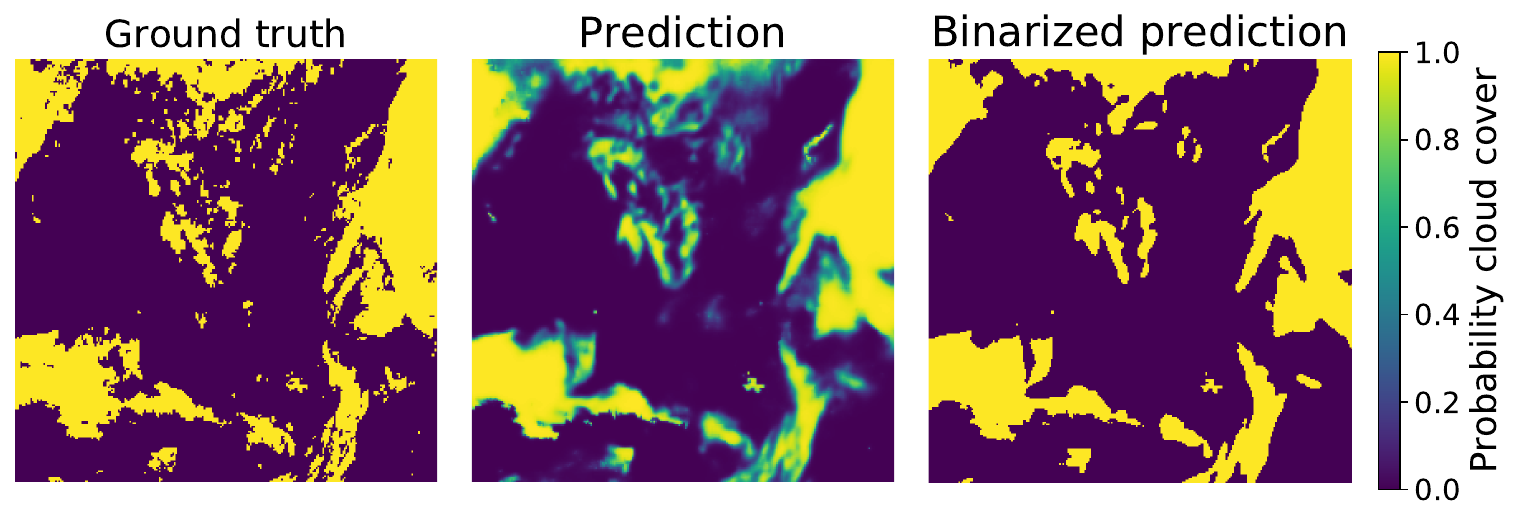}
\centering
\includegraphics[trim={0 0.1cm 0 0.1cm},clip,width=1.02\columnwidth]{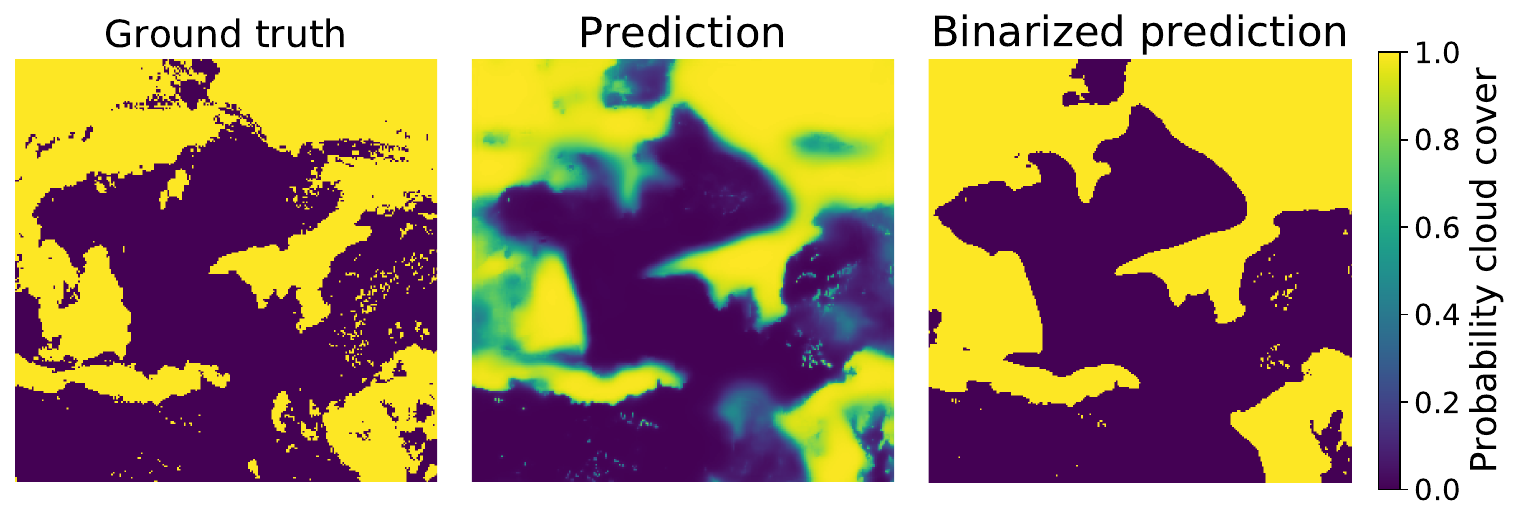}
\caption{Broad-UNet cloud cover prediction examples. The image above is predicted 30 mins ahead, and the image below is predicted 90 minutes ahead.}
\label{fig:cloudsImages}
\end{figure}

\section{Discussion}\label{sec:discussion}

\begin{figure*}[!htbp]
    \centering
    \includegraphics[scale=0.52]{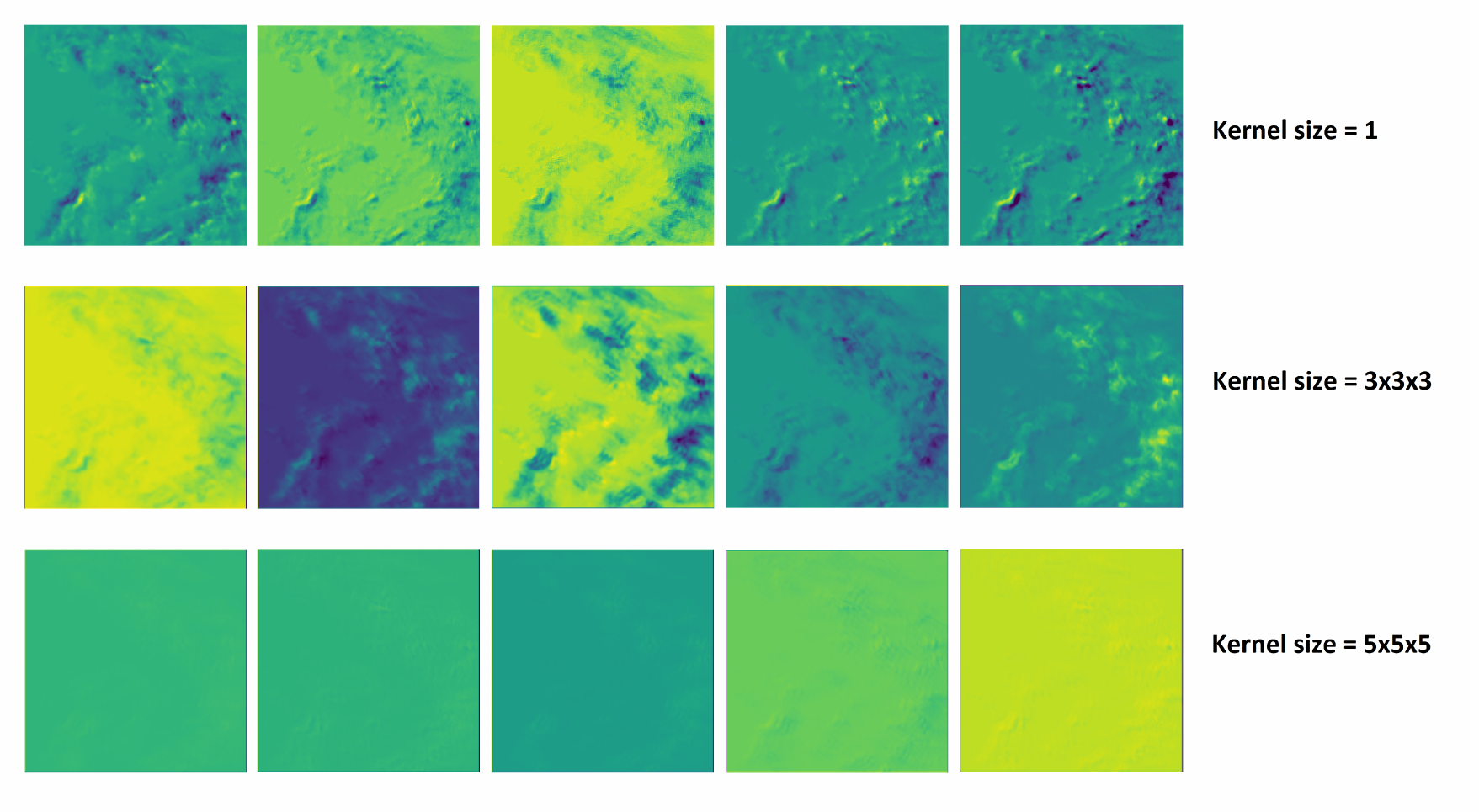}
    \caption{Feature maps outputted by different branches inside the first multi-scale feature convolutional block. Every row represents the output of a different branch. Next to each row, the kernel employed by the convolution in that branch is shown.}
    \label{fig:featsMaps}
\end{figure*}

From the obtained results, one can observe that the multi-scale feature learning allows the Broad-UNet to perform more precise predictions. This is thanks to the use of different convolutional filters in parallel. By combining convolutions with larger and smaller kernels, the model considers different amounts of information around the same region to generate the feature maps. Likewise, the inclusion of the ASPP module in the architecture allows the network to apply convolutions with diverse receptive fields at the same time. 

In the precipitation nowcasting task, we can observe an 11\% and an 8\% improvement with respect to the simple UNet for both datasets. In the cloud cover nowcasting task, the binarized predictions of the Broad-UNet are 7\% more accurate than the simple UNet for 15 minutes ahead predictions, and 1\% more accurate for 90 minutes ahead predictions. Since in the first nowcasting task, i.e. precipitation prediction, the model aims to perform a regression of each pixel with a wide range of values, achieving accurate forecasting or equivalently lower MSE values is more desirable. That is where the Broad-UNet shows more superior performance respect to the UNet. In the second nowcasting task, where the goal is to carry out a binary classification on each pixel, Broad-UNet performs slightly more accurate predictions than the UNet. While the immediate predictions (i.e. 15 and 30 mins ahead) are more precise, more distant predictions (more than 45 mins ahead) are comparable to UNet's predictions. Therefore, we can state that the wide building blocks of the Broad-UNet let the network to extract the spatial and short-term temporal information more accurately than the regular UNet. 

\begin{figure}[!htbp]
\centering
\includegraphics[width=0.44\columnwidth]{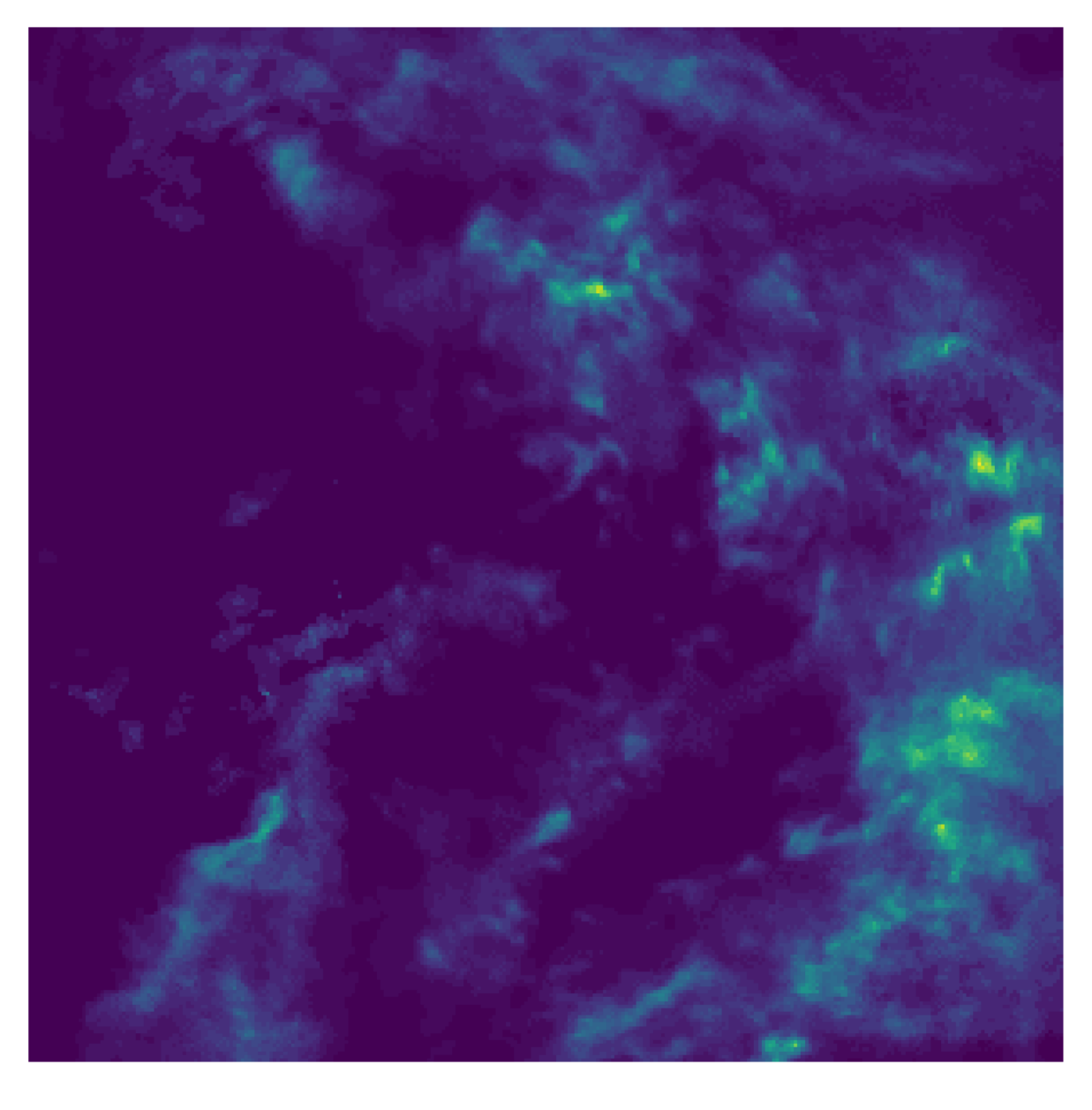}
\caption{Image fed to the network to generate the feature maps shown below. It belongs to the precipitation maps datasets, specifically to th 50\% of rain pixel dataset.}
\label{fig:origImgFeatsMaps}
\end{figure}

The learnt feature maps in different branches inside a convolutional block is shown in Fig. \ref{fig:featsMaps}. 
The chosen convolutional block is the first one so that the data doesn't have too abstract representation and is thus easier to interpret. The image fed to the network belongs to the precipitation maps dataset, and it is shown in Fig. \ref{fig:origImgFeatsMaps}. In Fig. \ref{fig:featsMaps}, the first row of the feature maps is the output of the convolutional branch with kernel size 1. The second row corresponds to the output of the branch with kernel size 3x3x3. Lastly, the third row corresponds to the output of the branch with kernel size 5x5x5. From Fig. \ref{fig:origImgFeatsMaps}, one can observe the differences between the features extracted in each branch. The convolutions with kernel size 1 seem to strengthen detailed differences in the image, and convolutions with kernel size 3x3x3 seem to accentuate differences between areas containing a high and a low rain concentration. In addition, convolutions with kernel size 5x5x5 seem to highlight regions with high rain concentration. 
\section{Conclusion}\label{sec:conclusion}
In this paper the Broad-UNet, an extension of the UNet architecture, is introduced for precipitation as well as cloud cover nowcasting. 
Thanks to the combination of the multi-scale feature convolutional block and the incorporation of the ASPP module, the proposed network is able to capture multi-scale information. In addition, the use of factorized kernels drastically reduces the number of parameters in the network compared to classical UNet model 
The performance of Broad-UNet is examined for addressing two nowcasting problems. The first problem consists of predicting precipitation maps 30 mins ahead. The second one consists of forecasting cloud cover 15 to 90 mins ahead. 
The obtained results suggest that the Broad-UNet extracts features more efficiently and therefore performs more accurate predictions in short-term nowcasting tasks compared to other tested UNet based models.  

\section*{Acknowledgment}
We would like to thank Léa Berthomier from Meteo France for providing the cloud cover dataset, as well as the required code to preprocess it.



\bibliography{Main}

\begin{thebibliography}{10}
\expandafter\ifx\csname url\endcsname\relax
  \def\url#1{\texttt{#1}}\fi
\expandafter\ifx\csname urlprefix\endcsname\relax\def\urlprefix{URL }\fi
\expandafter\ifx\csname href\endcsname\relax
  \def\href#1#2{#2} \def\path#1{#1}\fi

\bibitem{cogato2019extreme}
A.~Cogato, F.~Meggio, M.~De~Antoni~Migliorati, F.~Marinello, Extreme weather
  events in agriculture: A systematic review, Sustainability 11~(9) (2019)
  2547.

\bibitem{ivanov2019weather}
S.~Ivanov, P.~Ivanova, S.~Kuvshinkin, Weather conditions as a factor affecting
  the performance of modern powerful mining excavators, in: Journal of Physics:
  Conference Series, Vol. 1399, IOP Publishing, 2019, p. 044070.

\bibitem{senouci2018impact}
A.~Senouci, M.~Al-Abbasi, N.~N. Eldin, Impact of weather conditions on
  construction labour productivity in qatar, Middle East Journal of Management
  5~(1) (2018) 34--49.

\bibitem{sun2014use}
J.~Sun, M.~Xue, J.~W. Wilson, I.~Zawadzki, S.~P. Ballard, J.~Onvlee-Hooimeyer,
  P.~Joe, D.~M. Barker, P.-W. Li, B.~Golding, et~al., Use of nwp for nowcasting
  convective precipitation: Recent progress and challenges, Bulletin of the
  American Meteorological Society 95~(3) (2014) 409--426.

\bibitem{woo2017operational}
W.-c. Woo, W.-k. Wong, Operational application of optical flow techniques to
  radar-based rainfall nowcasting, Atmosphere 8~(3) (2017) 48.

\bibitem{shi2015convolutional}
X.~Shi, Z.~Chen, H.~Wang, D.-Y. Yeung, W.-K. Wong, W.-c. Woo, Convolutional
  lstm network: A machine learning approach for precipitation nowcasting,
  Advances in neural information processing systems 28 (2015) 802--810.

\bibitem{holmstrom2016machine}
M.~Holmstrom, D.~Liu, C.~Vo, Machine learning applied to weather forecasting,
  Stanford University (2016) 2--4.

\bibitem{grover2015deep}
A.~Grover, A.~Kapoor, E.~Horvitz, A deep hybrid model for weather forecasting,
  in: Proceedings of the 21th ACM SIGKDD International Conference on Knowledge
  Discovery and Data Mining, 2015, pp. 379--386.

\bibitem{faloutsos2019classical}
C.~Faloutsos, J.~Gasthaus, T.~Januschowski, Y.~Wang, Classical and contemporary
  approaches to big time series forecasting, in: Proceedings of the 2019
  International Conference on Management of Data, 2019, pp. 2042--2047.

\bibitem{webb2018deep}
S.~Webb, Deep learning for biology, Nature 554~(7693) (2018).

\bibitem{mehrkanoon2018deep}
S.~Mehrkanoon, J.~A.~K. Suykens, Deep hybrid neural-kernel networks using
  random fourier features, Neurocomputing 298 (2018) 46--54.

\bibitem{mehrkanoon2019deep}
S.~Mehrkanoon, Deep neural-kernel blocks, Neural Networks 116 (2019) 46--55.

\bibitem{mehrkanoon2015learning}
S.~Mehrkanoon, J.~A.~K. Suykens, Learning solutions to partial differential
  equations using ls-svm, Neurocomputing 159 (2015) 105--116.

\bibitem{mehrkanoon2019cross}
S.~Mehrkanoon, Cross-domain neural-kernel networks, Pattern Recognition Letters
  125 (2019) 474--480.

\bibitem{gamboa2017deep}
J.~C.~B. Gamboa, Deep learning for time-series analysis, arXiv preprint
  arXiv:1701.01887 (2017).

\bibitem{salman2015weather}
A.~G. Salman, B.~Kanigoro, Y.~Heryadi, Weather forecasting using deep learning
  techniques, in: 2015 international conference on advanced computer science
  and information systems (ICACSIS), IEEE, 2015, pp. 281--285.

\bibitem{coban2018neuro}
R.~Coban, I.~O. Aksu, Neuro-controller design by using the multifeedback layer
  neural network and the particle swarm optimization, Tehni{\v{c}}ki vjesnik
  25~(2) (2018) 437--444.

\bibitem{coban2013context}
R.~Coban, A context layered locally recurrent neural network for dynamic system
  identification, Engineering Applications of Artificial Intelligence 26~(1)
  (2013) 241--250.

\bibitem{voulodimos2018deep}
A.~Voulodimos, N.~Doulamis, A.~Doulamis, E.~Protopapadakis, Deep learning for
  computer vision: A brief review, Computational intelligence and neuroscience
  2018 (2018).

\bibitem{lu2007survey}
D.~Lu, Q.~Weng, A survey of image classification methods and techniques for
  improving classification performance, International journal of Remote sensing
  28~(5) (2007) 823--870.

\bibitem{goel2020state}
R.~Goel, A.~Sharma, R.~Kapoor, State-of-the-art object recognition techniques:
  A comparative study, in: Soft Computing: Theories and Applications, Springer,
  2020, pp. 925--932.

\bibitem{krizhevsky2017imagenet}
A.~Krizhevsky, I.~Sutskever, G.~E. Hinton, Imagenet classification with deep
  convolutional neural networks, Communications of the ACM 60~(6) (2017)
  84--90.

\bibitem{he2016deep}
K.~He, X.~Zhang, S.~Ren, J.~Sun, Deep residual learning for image recognition,
  in: Proceedings of the IEEE conference on computer vision and pattern
  recognition, 2016, pp. 770--778.

\bibitem{szegedy2015going}
C.~Szegedy, W.~Liu, Y.~Jia, P.~Sermanet, S.~Reed, D.~Anguelov, D.~Erhan,
  V.~Vanhoucke, A.~Rabinovich, Going deeper with convolutions, in: Proceedings
  of the IEEE conference on computer vision and pattern recognition, 2015, pp.
  1--9.

\bibitem{zhang2019light}
J.~Zhang, Y.~Xie, P.~Zhang, H.~Chen, Y.~Xia, C.~Shen, Light-weight hybrid
  convolutional network for liver tumor segmentation., in: IJCAI, 2019, pp.
  4271--4277.

\bibitem{berthomier2020cloud}
L.~Berthomier, B.~Pradel, L.~Perez, Cloud cover nowcasting with deep learning,
  arXiv preprint arXiv:2009.11577 (2020).

\bibitem{fernandez2020deep}
J.~G. Fern{\'a}ndez, I.~A. Abdellaoui, S.~Mehrkanoon, Deep coastal sea elements
  forecasting using u-net based models, arXiv preprint arXiv:2011.03303 (2020).

\bibitem{baldi2012autoencoders}
P.~Baldi, Autoencoders, unsupervised learning, and deep architectures, in:
  Proceedings of ICML workshop on unsupervised and transfer learning, 2012, pp.
  37--49.

\bibitem{lample2017unsupervised}
G.~Lample, A.~Conneau, L.~Denoyer, M.~Ranzato, Unsupervised machine translation
  using monolingual corpora only, arXiv preprint arXiv:1711.00043 (2017).

\bibitem{chung2016audio}
Y.-A. Chung, C.-C. Wu, C.-H. Shen, H.-Y. Lee, L.-S. Lee, Audio word2vec:
  Unsupervised learning of audio segment representations using
  sequence-to-sequence autoencoder, arXiv preprint arXiv:1603.00982 (2016).

\bibitem{ronneberger2015u}
O.~Ronneberger, P.~Fischer, T.~Brox, U-net: Convolutional networks for
  biomedical image segmentation, in: International Conference on Medical image
  computing and computer-assisted intervention, Springer, 2015, pp. 234--241.

\bibitem{trebing2021smaat}
K.~Trebing, T.~Sta\`{n}czyk, S.~Mehrkanoon, Smaat-unet: Precipitation
  nowcasting using a small attention-unet architecture, Pattern Recognition
  Letters 145 (2021) 178--186.

\bibitem{tao2017background}
Y.~Tao, P.~Palasek, Z.~Ling, I.~Patras, Background modelling based on
  generative unet, in: 2017 14th IEEE International Conference on Advanced
  Video and Signal Based Surveillance (AVSS), IEEE, 2017, pp. 1--6.

\bibitem{chen2017deeplab}
L.-C. Chen, G.~Papandreou, I.~Kokkinos, K.~Murphy, A.~L. Yuille, Deeplab:
  Semantic image segmentation with deep convolutional nets, atrous convolution,
  and fully connected crfs, IEEE transactions on pattern analysis and machine
  intelligence 40~(4) (2017) 834--848.

\bibitem{woo2018cbam}
S.~Woo, J.~Park, J.-Y. Lee, I.~S. Kweon, Cbam: Convolutional block attention
  module, in: Proceedings of the European conference on computer vision (ECCV),
  2018, pp. 3--19.

\bibitem{bowler2004development}
N.~E. Bowler, C.~E. Pierce, A.~Seed, Development of a precipitation nowcasting
  algorithm based upon optical flow techniques, Journal of Hydrology 288~(1-2)
  (2004) 74--91.

\bibitem{li2018subpixel}
L.~Li, Z.~He, S.~Chen, X.~Mai, A.~Zhang, B.~Hu, Z.~Li, X.~Tong, Subpixel-based
  precipitation nowcasting with the pyramid lucas--kanade optical flow
  technique, Atmosphere 9~(7) (2018) 260.

\bibitem{ayzel2019all}
G.~Ayzel, M.~Heistermann, A.~Sorokin, O.~Nikitin, O.~Lukyanova, All
  convolutional neural networks for radar-based precipitation nowcasting,
  Procedia Computer Science 150 (2019) 186--192.

\bibitem{agrawal2019machine}
S.~Agrawal, L.~Barrington, C.~Bromberg, J.~Burge, C.~Gazen, J.~Hickey, Machine
  learning for precipitation nowcasting from radar images, arXiv preprint
  arXiv:1912.12132 (2019).

\bibitem{shi2017deep}
X.~Shi, Z.~Gao, L.~Lausen, H.~Wang, D.-Y. Yeung, W.-k. Wong, W.-c. Woo, Deep
  learning for precipitation nowcasting: A benchmark and a new model, in:
  Advances in neural information processing systems, 2017, pp. 5617--5627.

\bibitem{lebedev2019precipitation}
V.~Lebedev, V.~Ivashkin, I.~Rudenko, A.~Ganshin, A.~Molchanov, S.~Ovcharenko,
  R.~Grokhovetskiy, I.~Bushmarinov, D.~Solomentsev, Precipitation nowcasting
  with satellite imagery, in: Proceedings of the 25th ACM SIGKDD International
  Conference on Knowledge Discovery \& Data Mining, 2019, pp. 2680--2688.

\bibitem{vaswani2017attention}
A.~Vaswani, N.~Shazeer, N.~Parmar, J.~Uszkoreit, L.~Jones, A.~N. Gomez,
  {\L}.~Kaiser, I.~Polosukhin, Attention is all you need, in: Advances in
  neural information processing systems, 2017, pp. 5998--6008.

\bibitem{szegedy2016inception}
C.~Szegedy, S.~Ioffe, V.~Vanhoucke, A.~Alemi, Inception-v4, inception-resnet
  and the impact of residual connections on learning, arXiv preprint
  arXiv:1602.07261 (2016).

\bibitem{yang2019asymmetric}
H.~Yang, C.~Yuan, B.~Li, Y.~Du, J.~Xing, W.~Hu, S.~J. Maybank, Asymmetric 3d
  convolutional neural networks for action recognition, Pattern recognition 85
  (2019) 1--12.

\bibitem{KNMI}
R.~N.~M. Institute, \href{https://www.knmi.nl/home}{Royal netherlands
  meteorological institute} (2020).
\newline\urlprefix\url{https://www.knmi.nl/home}

\bibitem{clouds}
EUMETSAT,
  \href{https://navigator.eumetsat.int/product/EO:EUM:DAT:MSG:GNWCCT}{Geostationary
  nowcasting cloud type - msg - 0 degree} (2020).
\newline\urlprefix\url{https://navigator.eumetsat.int/product/EO:EUM:DAT:MSG:GNWCCT}

\end{thebibliography}

\end{document}